\def\year{2022}\relax
\documentclass[letterpaper]{article} 
\usepackage{aaai22}  
\usepackage{times}  
\usepackage{helvet}  
\usepackage{courier}  
\usepackage[hyphens]{url}  
\usepackage{graphicx} 
\usepackage{natbib}  
\usepackage{caption} 
\DeclareCaptionStyle{ruled}{labelfont=normalfont,labelsep=colon,strut=off} 
\usepackage{algorithm}
\usepackage{algorithmic}
\usepackage{booktabs}
\usepackage{amsmath}
\usepackage{amssymb}
\usepackage{multirow}

\usepackage{xcolor}

\usepackage{newfloat}
\usepackage{listings}
\floatstyle{ruled}
\newfloat{listing}{tb}{lst}{}
\floatname{listing}{Listing}
\nocopyright 

\title{Studying Lobby Influence in the European Parliament}
\author{
Aswin Suresh,
Lazar Radojevic\equalcontrib,
Francesco Salvi\equalcontrib, \\
Antoine Magron,
Victor Kristof,
Matthias Grossglauser
}
\affiliations{
EPFL \\
Switzerland \\
firstname.lastname@epfl.ch
}

\usepackage{bibentry}

\begin{document}

\maketitle

\begin{abstract}
We present a method based on natural language processing (NLP), for studying the influence of interest groups (lobbies) in the law-making process in the European Parliament (EP).
We collect and analyze novel datasets of lobbies' position papers and speeches made by members of the EP (MEPs).
By comparing these texts on the basis of semantic similarity and entailment, we are able to discover interpretable links between MEPs and lobbies.
In the absence of a ground-truth dataset of such links, we perform an indirect validation by comparing the discovered links with a dataset, which we curate, of retweet links between MEPs and lobbies, and with the publicly disclosed meetings of MEPs.
Our best method achieves an AUC score of 0.77 and performs significantly better than several baselines.
Moreover, an aggregate analysis of the discovered links, between groups of related lobbies and political groups of MEPs, correspond to the expectations from the ideology of the groups (e.g., center-left groups are associated with social causes).
We believe that this work, which encompasses the methodology, datasets, and results, is a step towards enhancing the transparency of the intricate decision-making processes within democratic institutions.
\end{abstract}

\section{Introduction}

The transparency of decision-making is of central importance for the legitimacy of democratic institutions such as parliaments.
The influence of interest groups (lobbies) on parliamentarians and the potential for a resultant subversion of the power of the electorate to determine policy have led to demands from groups, such as \citet{transparency2023mission}, for effective rules and systems to increase transparency.
The emergence of several open government initiatives around the world \citep{switzerland2021open,eu2021data,open2018barack} is in part a response to such demands.

The EU Transparency Register (TR) \citep{eu2011transparency} is one such initiative that provides a tool for EU citizens to explore the influence of interest groups in the European Parliament (EP). 
Any organization that seeks to influence EU policy, with a few notable exceptions, needs to register with the TR before meeting with parliamentarians.
The organizations are asked to disclose information such as their address, website, financial information, and goals.

However, the EU TR has several limitations.
The disclosure of most of the information is voluntary and there is little oversight.
It is difficult to obtain information regarding which members of the EP (MEPs) or laws are targeted (and by which particular lobbies) and to know the lobbies' positions on specific policies.

There have been several studies conducted by the political science community on EU lobbying.
\citet{bouwen2003theoretical} develop a theoretical framework to explain the access of business lobbies to the EP and empirically test their hypotheses by means of interviews.
\citet{rasmussen2015battle} study the influence of business groups in shaping policy at the EP, focusing on a few dossiers and using interview data.
\citet{tarrant2022big} perform a case study of the influence of Big Tech on the Digital Markets Act in the EU.
These studies all focus either on a single policy issue or on a small set of issues, and/or they are limited in terms of sample size as they employ less scalable methodologies such as manual examination of position papers and individual interviews.

One exception is a study by \citet{ibenskas2021legislators}. They analyze the Twitter follower network of a large number of MEPs and lobbies from the TR, with respect to the MEP's nationality and committee memberships and lobbies' self-reported interests in the TR.
However, they do not analyze the textual content of MEPs' speeches and amendments and the lobbies' position papers, which would be instrumental for uncovering convergence on specific policy issues beyond the broad interest areas mentioned in the TR. 

Therefore, there is a need for automated approaches for studying lobbying in a comprehensive manner, with the help of rich publically available textual resources and by using modern tools developed by the NLP community.
In particular, self-supervised algorithms for text representation and computing text similarity and entailment \citep{reimers2019sentence} are promising for identifying interesting patterns.
A major challenge faced by such automated approaches, even unsupervised ones, is the lack of ground-truth data for validation.
As far as we are aware, there exists no large database of verified MEP-Lobby links, let alone one annotated for relevant policy positions.

In this work, we present an NLP-based approach for automatically discovering potential links between a large number of MEPs and lobbies, by comparing the text in publicly available documents where they express their views on policy issues.
We define a link between an MEP and a lobby as a convergence of views between them on some policy issue.
To the best of our knowledge, such an approach has not been explored in prior work.
We focus on the eighth term of the EP (2014-2019), as it was the last complete term that was not disrupted due to the pandemic.
In the absence of ground-truth data, we perform an indirect validation by comparing the discovered links to a dataset we curate of retweet links between MEPs and lobbies.

We use the retweet network instead of the follower network studied by \citet{ibenskas2021legislators}, because retweets typically occur as a result of the agreement of particular views between the MEP and lobby, in contrast to `follows' that can result from a general interest in knowing more about a topic or person \citep{metaxas2015retweets}.  
Moreover, timestamps for retweets are publicly available, which allows us to collect more relevant data for the eighth term. 

Since 2019, it has been mandatory for MEPs in certain key positions (such as reporters of parliamentary committees) to publish their meetings with lobby groups \citep{europarl2019mepmeetings}.
We use this data as an additional source of validation, although it only covers the subset of the MEPs from the eighth term who were re-elected in the ninth term.

Our methods are also designed to be interpretable - we can obtain the specific set of MEP speeches and lobby documents that match for an MEP-Lobby pair, thus enabling manual validation of discovered links by users.

This is particularly important since we do not claim that the presence of a discovered link between a particular MEP and lobby group necessarily implies that the MEP was influenced, duly or unduly, by the lobby. 
Rather, it means that the views of the MEP and lobby are probably similar on an issue that is referenced by the matched texts. 
This similarity could possibly be the result of influence, but such a claim needs to be validated further by the user by a careful examination of the matched texts for their similarity in addition to relevant contextual information.

In this paper, to avoid any harm to the reputation of MEPs through showing the spurious links that could result from inadvertent errors in our interpretation, we restrict ourselves to an aggregate analysis instead of showing individual MEP-lobby links.

The paper is structured as follows.
In Section \ref{sec:lb_datasets}, we describe the datasets that we curate and use.
In Section \ref{sec:lb_methods}, we describe the different methods that we experiment with for discovering links.
We evaluate the methods in Section \ref{sec:lb_evaluation} and interpret them in Section \ref{sec:lb_interpretation}.
We conclude the paper in Section \ref{sec:lb_conclusion}.

\section{Datasets}
\label{sec:lb_datasets}
We curate several novel datasets for our study.
To obtain the policy positions of lobbies, we curate a dataset of position papers (Section \ref{sec:lobbypp}).
The views of the MEPs are obtained through a dataset of their plenary speeches (Section \ref{sec:mepspeeches}) and proposed amendments (Section \ref{sec:mepamendments}).
For validation, we use a dataset of MEP-lobby retweet links (Section \ref{sec:meplobbyretweet}) and meetings (Section \ref{sec:meplobbymeetings}). 

\subsection{Lobbies}
\label{sec:lobbypp}

\begin{figure*}[h]
    \centering
    \includegraphics[trim=0cm 9.5cm 2cm 5.5cm, clip, width=0.9\textwidth]{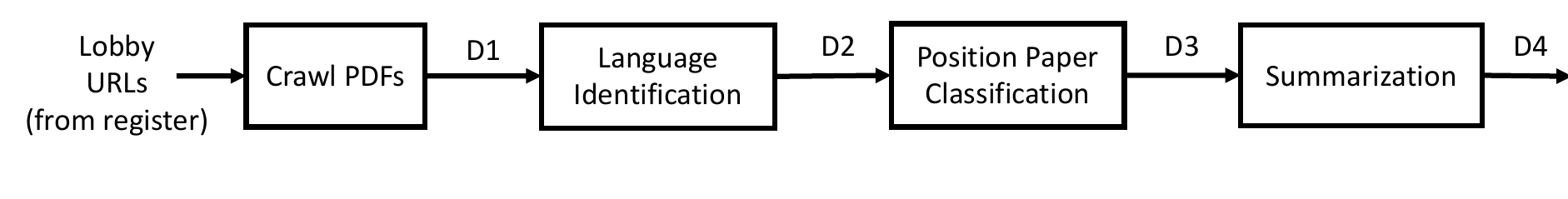}
    \caption{Data collection pipeline for lobbies. $D1$ contains all crawled PDF documents, $D2$ contains all English documents in $D1$, $D3$ contains the documents in $D2$ classified as position papers, and $D4$ contains the summaries of the documents in $D3$.}
    \label{fig:pipelinelobbyppdataset}
\end{figure*}

Our data collection pipeline for lobbies is given in Figure \ref{fig:pipelinelobbyppdataset}.
The versions of the data at different stages of the pipeline are labeled as $D1$, $D2$, and so on.
Information on the size of these datasets is given in Table \ref{tab:statslobbyppdataset}.
We now describe the steps in the pipeline.

\begin{table}[h]
\centering
\caption{Lobby datasets}
\label{tab:statslobbyppdataset}
\begin{tabular}{@{}llll@{}}
\toprule
   & $D1$ & $D2$ & $D3, D4$ \\ \midrule
Documents & 766,437  & 373,216 & 48,970 \\
Lobbies   & 4,230   & 3,965 & 2,558 \\ \bottomrule
\end{tabular}
\end{table}

\subsubsection{Crawling and Language Identification}
We focus on the lobbies that were on the EU TR under the heads of \textit{Trade and Business Associations}, \textit{Trade Unions and Professional Associaions} and \textit{Non-Governmental Organisations}, as of October 2020; this is a total of 5,461 lobbies.
Although some other categories like \textit{Companies and Groups} are also influential, we do not include them because they are mostly represented by associations that they are part of and rarely publish position papers of their own.

We obtain the URLs of the lobby websites from the TR and crawl publicly available PDF documents from them to obtain an initial dataset $D1$.
We parallelize the crawling by using HTCondor \citep{htc2023htc} on a cluster of 300 nodes with maximum limits of 250 MB of text and 5 hours of crawling per website and are able to crawl all PDFs in nearly 70\% of the lobby websites in about four days.
Spending several hours per website enables us to keep a sufficient interval between consecutive HTTP requests (similar to that of a human user) so that the functioning of the websites is not adversely affected.
We extract and store only the text from the PDF documents to keep storage costs manageable.

To identify the languages in the dataset, we use the Fasttext language identification model \citep{joulin2016langid, joulin2016fasttextzip}.
As nearly half (48.7\%) of all the documents are in English and other languages appear in much smaller percentages, we keep only the English documents ($D2$) to simplify the rest of the analysis.

\subsubsection{Position Paper Classification}
A large majority of the PDFs do not contain significant information about lobby policy positions, including documents such as product brochures, user manuals, technical documentation, forms, etc.
By manually labeling 200 randomly sampled PDFs\footnote{Two of the authors independently performed the labeling. Cohen's $\kappa$ was 0.4, indicating fair to moderate agreement. Disagreements were subsequently resolved by discussion.}, we estimate the proportion of PDFs that contain policy positions to be approximately 25\%.
In order to reduce noise in the data and to enable us to apply methods that are more performant but less scalable, we classify the PDFs into \textit{position papers} and \textit{other documents} and work with those classified as position papers.

We train a weakly supervised logistic regression model by using TF-IDF features for this task and by using the presence of the word `position' in the URL as the label.
On the manually labeled validation set of 200 PDFs, the model achieves a precision of 95\% and a recall of 39\% in identifying position papers.
The most predictive words include \textit{position}, \textit{should}, \textit{strongly}, etc. and are indeed likely to be present in texts articulating positions.
We then apply the classifier on all PDFs in $D2$, and keep those that are classified as position papers to obtain $D3$.

\subsubsection{Summarization}
\label{sec:summarization}
Many of the documents are quite long (greater than 1,000 words) and cannot be encoded fully by pre-trained encoders such as SentenceBERT \citep{reimers2019sentence}.
They also typically contain information, such as technical details, that is not relevant for matching with MEP speeches.
Hence, we summarize the documents into three-to-four sentences that capture the main ideas expressed. 
This also makes the interpretation of matched document-speech pairs easier.

We experiment with various state-of-the-art pre-trained summarization models, namely T5 \citep{raffel2020t5} and BART \citep{lewis2020bart}, and Large Language Models (LLMs) including OpenAI's \texttt{gpt-3.5-turbo} (the model behind ChatGPT) \citep{openai2023chatgptapi} and LMSYS's Vicuna-7B-v1.5 \citep{zheng2023judging} (a LLaMA-2 \citep{touvron2023llama2} model fine-tuned on ChatGPT conversations).
We manually compare five random summaries generated by each model and find that ChatGPT and Vicuna generate coherent summaries that capture the most salient points expressed in the document, while T5 and BART omit important information and generate disconnected sentences that are almost the same as those in the original document.

We thus generate the dataset $D4$ that contains the LLM-generated summaries of documents in $D3$.
We summarize only the documents in $D3$, despite the low recall of position paper classification due to the cost constraints of using LLMs.

\subsubsection{Lobby Clustering}
Individual lobbies are so numerous and specialized that it is difficult to see interpretable patterns, even after a successful MEP-Lobby matching.
We, therefore, cluster the lobbies into relatively homogenous groups by using the description of their goals in the EU TR.
We first convert these descriptions to short phrases (3-4 words) by using ChatGPT and cluster the phrases by using K-Means\footnote{We use $K=100$ as it gives mostly coherent clusters with minimal duplicates.} after embedding them using SentenceBERT.
The clusters are mostly straightforward to interpret, although some of them contain a few unrelated lobbies.
Some clusters, related particularly to energy, include both renewable energy companies and fossil-fuel companies.
This is probably because some of the fossil-fuel companies are undergoing a renewables transition and emphasize this in their goal statements in the TR.

Finally, we ask ChatGPT to name each of the clusters, based on the short phrase descriptions of the lobbies in each of them.
Most of the names are highly representative and specific, but a few of them are too generic (e.g., `Interest groups in the EU');  we then manually correct the overly generic ones to make them more specific.
The list of the top three lobby clusters with the most position papers is given in Table \ref{tab:toplobbyclusters} with a couple of examples of lobbies that are in each cluster. 
A complete list of lobby clusters that we refer to in this paper, with example lobbies, is given in Appendix A.

\begin{table}[h]
\centering
\caption{Top three lobby clusters by number of position papers. All three have about 1,400 papers each.}
\label{tab:toplobbyclusters}
\begin{tabular}{@{}ll@{}}
\toprule
\textbf{Lobby Cluster} & \textbf{Example Lobbies}                                                      \\ \midrule
Manufacturing          & \begin{tabular}[c]{@{}l@{}} orgalim.eu \\ glassforeurope.com \end{tabular}       \\ \midrule
Renewable Energy      & \begin{tabular}[c]{@{}l@{}} solarpowereurope.org\\ windeurope.org\end{tabular} \\ \midrule
Business               & \begin{tabular}[c]{@{}l@{}}enterprisealliance.eu\\ smeeurope.eu\end{tabular} \\ \bottomrule
\end{tabular}
\end{table}

\subsection{MEPs}
Data on MEPs' policy positions are obtained from two sources: their speeches in the plenary sessions of the EP, and the law amendments that they propose within parliamentary committees.
We describe each of them in the following sections.

\subsubsection{Speeches}
\label{sec:mepspeeches}
We scrape all plenary speeches of the eighth term from the EP website (51,432 in total), spoken by 849 MEPs (and a few non-members).
The speeches are organized into 1,471 debates with titles; each debate is about a specific law or policy issue.
For the speeches made by MEPs, we scrape the official EP ID of the MEP, which we use to query the Parltrack database \citep{parltrack2023parltrack} to obtain additional information about the MEP, such as their name, nationality, party, etc.

Similar to the case for lobbies, it is easier to find patterns if we analyze the links to lobbies for \textit{groups} of MEPs rather than individuals.
MEPs are naturally grouped according to their ideology into nine political groups.
The European People's Party (EPP, center-right) and the Socialists and Democrats (S\&D, center-left) are the two largest groups.

To quantify the ideological position of the groups, we use data from the Chapel Hill Expert Survey (CHES) \citep{jolly2022chapel}, where political scientists have scored every party on a numerical ideological scale ranging from zero (extreme left) to ten (extreme right).
In addition to the general left-right ideology (which are referred to simply as `Ideology'), the survey also contains scores for more fine-grained aspects of ideology such as views on how to manage the \textit{economy} (state control vs. free market), views on \textit{social} issues (libertarian vs. traditional/authoritarian), and views on \textit{EU} integration (anti-EU vs. pro-EU)\footnote{The CHES codebook refers to these scores as LRGEN, LRECON, GALTAN, and EU\_POSITION}.
We aggregate the party-level data from CHES to get the scores for the political groups\footnote{We take the weighted average of party scores with weights being the size of each party in the group.}. 
The positions of the nine EP groups are given in Table \ref{tab:politicalgroups}.

\begin{table*}[h]
\centering
\caption{Political groups and ideology scores, sorted by general left-right ideology.}
\label{tab:politicalgroups}
\begin{tabular}{@{}p{0.35\textwidth}llllll@{}}
\toprule
\textbf{Group name} & \textbf{Acronym} & \textbf{Ideo} & \textbf{Econ} & \textbf{Soc} & \textbf{EU}\\ \midrule
Confederal Group of the European United Left - Nordic Green Left & GUE/NGL & 1.65 & 1.39 & 3.31 & 3.49\\ \midrule
Group of the Greens/European Free Alliance & Greens/EFA & 3.21 & 3.22 & 2.21 & 5.61 \\ \midrule
Group of the Progressive Alliance of Socialists and Democrats in the European Parliament & S\&D & 3.83 & 3.90 & 3.83 & 6.18 \\ \midrule
Group of the Alliance of Liberals and Democrats for Europe & ALDE & 6.09 & 6.70 & 4.00 & 6.05 \\ \midrule
Europe of Freedom and Direct Democracy Group & EFDD & 6.55 & 5.43 & 5.63 & 1.40\\ \midrule
Group of the European People's Party (Christian Democrats) & EPP & 6.69 & 6.32 & 6.38 & 5.89 \\ \midrule
European Conservatives and Reformists Group & ECR & 7.21 & 5.90 & 7.28 & 3.33 \\ \midrule
Europe of Nations and Freedom Group & ENF & 9.32 & 6.14 & 8.89 & 1.31\\ \midrule
Non-Attached Members & NI & 9.76 & 4.06 & 9.54 & 1.18
\end{tabular}
\end{table*}

The speeches for the eighth term are available only in the original language of the speaker, unlike in earlier terms where the EP provided translated versions in all official EU languages, including English.
Hence, we automatically translate all the non-English speeches to English by using the open-source OPUS-MT models \citep{tiedemann2020opusmt} provided in the EasyNMT package \citep{reimers2022easynmt}.
Apart from the dataset of full speeches, we also generate a dataset of the speech summaries by using LLMs as in Section \ref{sec:summarization}.
We use the summarized speeches when matching with the summaries of the lobbies' position papers ($D4$), and the full speeches when matching with the full lobby documents ($D2$ and $D3$).

\subsubsection{Amendments}
\label{sec:mepamendments}
We use the law amendments dataset released by \citet{kristof2021war}, which contains 104,996 amendments proposed by MEPs in the eighth term on 347 laws identified by their titles.
We input the old and new versions of the law articles changed by the amendment to an LLM (either ChatGPT or Vicuna), along with the law title, and ask it to generate a possible sentence for the position paper of a lobby that would like to get this amendment accepted.
We expect to be able to match the sentence to the lobby summaries ($D4$) generated in Section \ref{sec:summarization}.

We find that the LLMs generate a concise summary of the amendment's effect on the law, adding that they (by impersonating a lobbyist) would like such a change to be effected. 
Short but significant changes to the law are correctly interpreted by the model, such as the change from \textit{shall} to \textit{should} being a change from a mandatory requirement to a recommendation.
However, the model tends to ``hallucinate'' when there is insufficient context, such as in the case of entire articles being deleted or new ones being added. 
Therefore, we restrict this procedure to generate summaries exclusively for the 88,853 amendments that only modify existing articles without deleting them entirely.

\subsection{Validation Datasets}
For validating the discovered links, we curate a dataset of retweet links and a dataset of MEP-Lobby meetings.
We describe them in the following sections.

\subsubsection{MEP-Lobby Retweet Links}
\label{sec:meplobbyretweet}
We obtain the Twitter handles of MEPs from multiple sources including official profile pages on the EP website, the Parltrack database, other third-party databases, and manual search.
We were able to obtain handles for 666 MEPs.
We collect handles of the lobbies with position papers by scraping their homepages for `Follow us on Twitter' links, and obtain 1,676 handles.
We see that, indeed, most of the MEPs and Lobbies have a presence on Twitter.

Once we have the handles, we use the Full Archive Search endpoint of the Twitter API\footnote{The Twitter API changed recently and no longer provides this level of access for free.} \citep{twitter2023twitterapi} to retrieve the content and metadata of all their public tweets during the period of the eighth term.
We then identify the tweets of an MEP (resp. lobby) that are `pure' retweets (without any added original content hence less likely to indicate disagreement) and check if the referenced tweet is from a lobby (resp. MEP).
We consider that there is an (undirected) retweet link between an MEP-Lobby pair if either the MEP or the lobby has retweeted the other at least once, which leaves us with 8,754 links. 

\subsubsection{MEP-Lobby Meeting Links}
\label{sec:meplobbymeetings}
Data on meetings between MEPs and lobbies since the beginning of the ninth EP term (2019-2024) are available from the Integrity Watch Data Hub \citep{integritywatch}. 
Integrity Watch monitors and collects meeting information from the European Parliament website. 
Every meeting includes an MEP identified by the EP ID and a list of lobby names or acronyms.
We match the lobby names to our data from the register using fuzzy string matching, thus enabling us to establish 1,365 links between 125 MEPs from the eighth term (who were re-elected in the 9th term) and 565 lobbies.

\section{Methods}
\label{sec:lb_methods}
Here, we describe the framework and methods we use to discover links between MEPs and lobbies.
Let $\mathcal{M}$ denote a set of MEPs and $\mathcal{L}$ denote a set of lobbies.
We assume that an MEP $m \in \mathcal{M}$ and a lobby $l \in \mathcal{L}$ have a link with some probability $P(m,l)$.
One possible approach to discovering links is to estimate this probability directly.
However, this is difficult as we do not have a ground-truth dataset of links on which to train a probabilistic model.
Without such data, we can make only \textit{relative} assessments of $P(m,l)$, based on information about the similarity of views between $m$ and $l$.
Thus, we can say that $P(m_1,l_1) > P(m_2,l_2)$ if the similarity of views for the pair $(m_1,l_1)$ is higher than that for the pair $(m_2,l_2)$.

Hence, we adopt the following framework.
Given an MEP $m \in \mathcal{M}$, and a lobby $l \in \mathcal{L}$, the goal of our methods is to compute an association score $A(m,l) \in \mathbb{R}$ such that
\begin{align}
    & \nonumber A(m_1,l_1) > A(m_2, l_2) \\ 
    & \nonumber \quad \iff P(m_1,l_1) > P(m_2,l_2) \\ 
    &\quad \quad \forall m_1,m_2 \in \mathcal{M},~~\forall l_1,l_2 \in \mathcal{L}.
\end{align}
The methods differ in how $A(m,l)$ is computed.
For methods using texts, we use $\mathcal{S}_m$ to refer to the documents produced by $m$ and $\mathcal{D}_l$ for the documents produced by $l$.

\subsection{Baselines}
We first describe the baselines. 
The goal of comparing our models to these baselines is to check if the content of the texts provides non-trivial information about the MEP-Lobby association.

\subsubsection{Random}
This is the simplest baseline where we have $A(m,l) \sim \text{Uniform}(0, 1)$.

\subsubsection{Prolificacy (Pr)}
This baseline is based on the intuition that the MEPs and lobbies that are more prolific and generate more texts are more likely to have a link between them.
Hence, for this baseline, we define $A(m,l) = \lvert \mathcal{S}_m \rvert  \times \lvert \mathcal{D}_l \rvert.$

\subsubsection{Nationality (Nat)}
Prior work suggests that there is a strong tendency for MEPs to associate with lobbies from the same EU member state \citep{ibenskas2021legislators}.
We therefore include a baseline where $A(m,l)=1$ if $m$ and $l$ are from the same member state and $A(m,l)=0$ otherwise.

\subsection{Text-Based Methods}
Here, we describe our methods that use the content of the texts in $\mathcal{S}_m$ and $\mathcal{D}_l$.

\subsubsection{Text Classification (Class)}

We train a fastText (supervised) classifier to predict whether a given text was generated by a particular lobby.
We use the sentences in the lobby dataset $D2$, creating a 80-20\% train-test data split. We train the classifier for $10$ epochs with a one-versus-all loss, a learning rate of $0.2$, and word n-grams of length up to $2$: such hyperparameters were selected as the best performing in a grid search over learning rate and loss. Independent linear classifiers are trained for each lobby, but they share the same embedding layer, which enables the model to scale to a large number of classes while having limited data for each class. 
The linear structure allows interpretability; the top predictive words for some lobbies are given in Table \ref{tab:fasttexttopwords}.
We see that these clearly reflect the areas of work of the lobbies.

\begin{table*}[h]
  \centering
        \addtolength{\tabcolsep}{-3pt}
        \begin{tabular}{c|ccccc}
        \toprule
        amnesty.eu & executions & detainee & occupants & assurances & reassignment \\
        businesseurope.eu & globalisation & kyoto & relocation & lisbon & wto \\
        caneurope.org & climate & warming & fossil & coal & allowances \\
        fuelseurope.eu & refineries & refinery & gasoline & fuels & cis \\
        ficpi.org & invention & trademarks & patent & practitioner & attorneys \\
        orgalim.eu & manufacturers & machines & engineering & doc & counterfeiting \\
        \bottomrule
    \end{tabular}%
    \addtolength{\tabcolsep}{3pt}
    \caption{Top predictive words for some prominent lobbies}
  \label{tab:fasttexttopwords}%
\end{table*}%

Once the classifier is trained, we compute 
\begin{equation}
A(m,l) = \frac{1}{\lvert \mathcal{S}_m \rvert} \sum_{s \in \mathcal{S}_m} P(l|s),     
\end{equation}
where $P(l|s)$ is the probability that lobby $l$ generated the text $s$, according to the trained classifier.

\subsubsection{Semantic Similarity (SS)}
\label{sec:semsim}
In this method, we first convert the texts in $\mathcal{S}_m$ and $\mathcal{D}_l$ to vector representations that capture their meaning.
The cosine similarity between these vectors gives a measure of semantic similarity between the texts.

We use the pre-trained \texttt{all-MiniLM-L6-v2} model from SentenceBERT to obtain 384-dimensional vector representations for the texts.
We then compute

\begin{equation}
A(m,l) = \max_{s \in \mathcal{S}_m, d \in \mathcal{D}_l} \mathbf{v_s}^T \mathbf{v_d},
\end{equation}

where $\mathbf{v_s}$ and $\mathbf{v_d}$ are the vector representations of texts $s$ and $d$ respectively, normalized to unit norm.

If the whole text fits within the maximum sequence length for the SentenceBERT model (256 tokens), it is encoded into a vector directly.
This is the case for summary texts.
If the text is too large to fit, we separate it into individual sentences and take the normalized sum of the sentence encodings.

\subsubsection{Entailment (Ent)}
One issue with SS is that there exist cases where two texts contradict each other, but they still have high semantic similarity based on their vector representations. 
This can cause false positives in the discovered links. 
For instance, an MEP's speech about increasing a specific tax could be matched with a lobby's position paper advocating for a reduction of the same tax.
One reason for this is that the fixed-length vector representation might not always have enough information to process negations.

In order to reduce such cases, we use a cross encoder model pre-trained on natural language inference (NLI) data, including SNLI and MultiNLI.
We use, in particular,  the \texttt{cross-encoder/nli-deberta-v3-base} model from SentenceBERT.
Given a pair of texts $(s,d)$, this model is trained to output whether $s$ contradicts $d$, $s$ entails $d$, or neither.

As texts from an MEP speech and lobby document are usually less similar than a pair of premise and hypothesis from NLI, the model assigns the highest probability to \textit{neither} for most of the text pairs.
However, we can identify probable contradictions, especially for highly similar pairs, by checking if the probability it assigns to \textit{contradiction} ($P(con)$) is greater than that for \textit{entailment} ($P(ent)$).

We then compute
\begin{align}
\nonumber A(m,l) = &\max_{s \in \mathcal{S}_m, d \in \mathcal{D}_l} \mathbf{v_s}^T \mathbf{v_d}, \\
& \text{s.t}~~p_{ent}(s,d) > p_{con}(s,d).  
\end{align}

\section{Evaluation}
\label{sec:lb_evaluation}

We evaluate our methods on both the retweet links and meetings datasets.
We use the area under the receiver operating characteristic (ROC) curve (AUC), as our metric because it is independent of the choice of a threshold for $A(m,l)$.
We are mostly interested in the low false positive rate (FPR) regime of the ROC as we expect the MEP-Lobby influence network to be sparse.
Hence, we also compute the partial AUC (pAUC) for the FPR $<$ 0.05 region.

\begin{figure}[h!]
  \centering
  \includegraphics[width=.9\columnwidth]{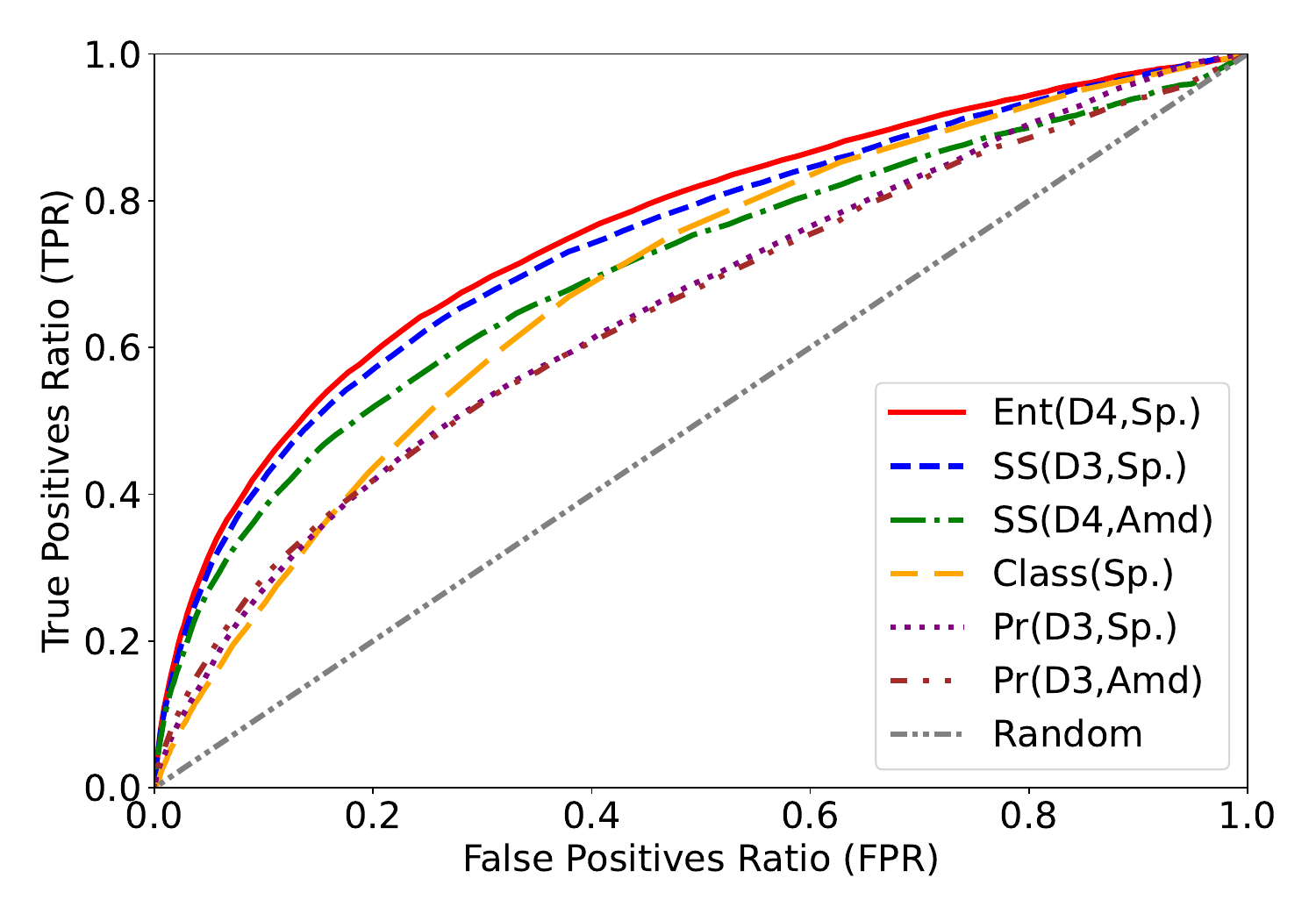} \\
  \includegraphics[width=.9\columnwidth]{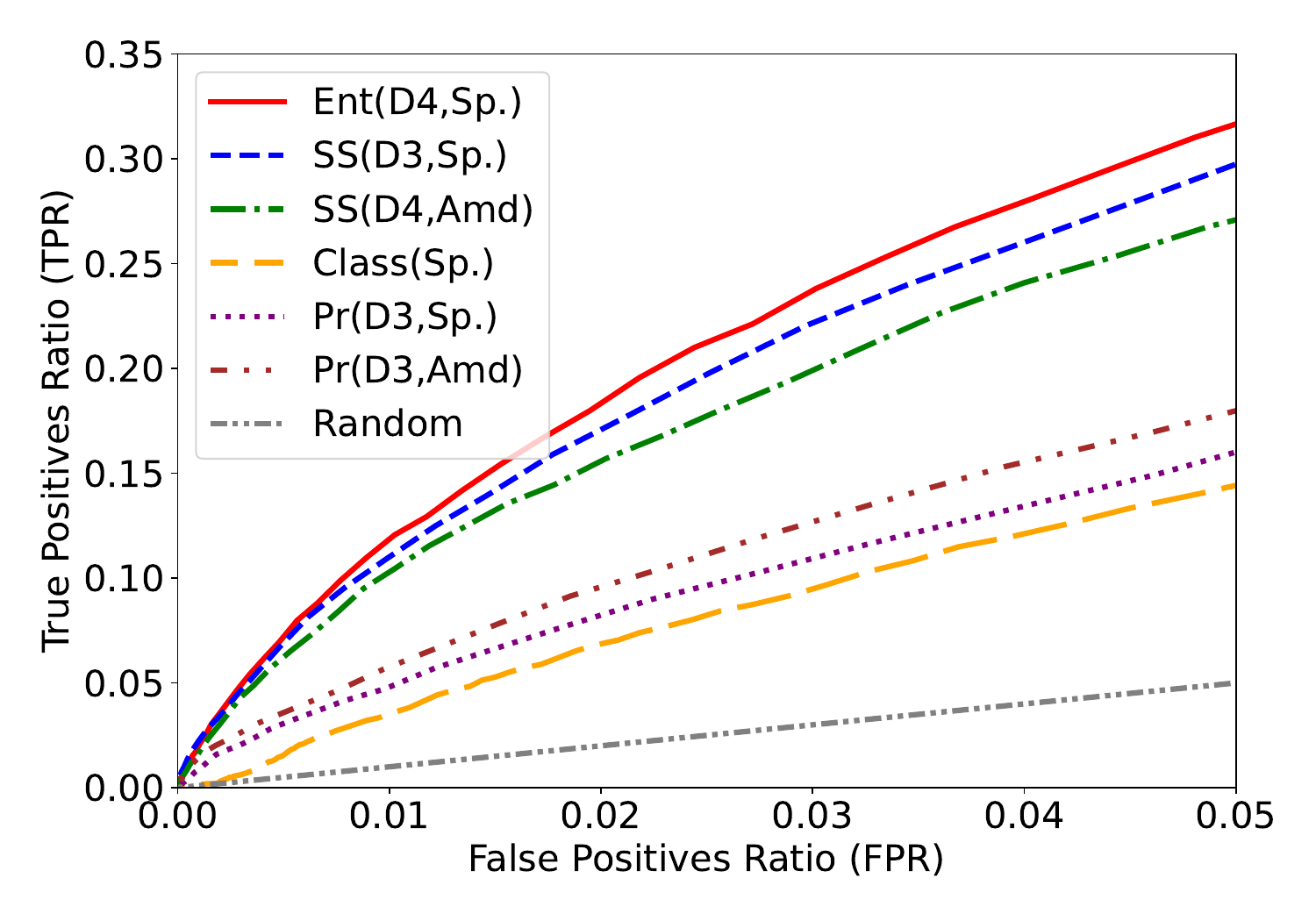}
  \caption{ROC curves for the Retweet dataset - Full (Top) and FPR$\leq$0.05 region (Bottom)}
  \label{fig:twitterroccurves}
\end{figure}

\begin{figure}[h!]
  \centering
  \includegraphics[width=.9\columnwidth]{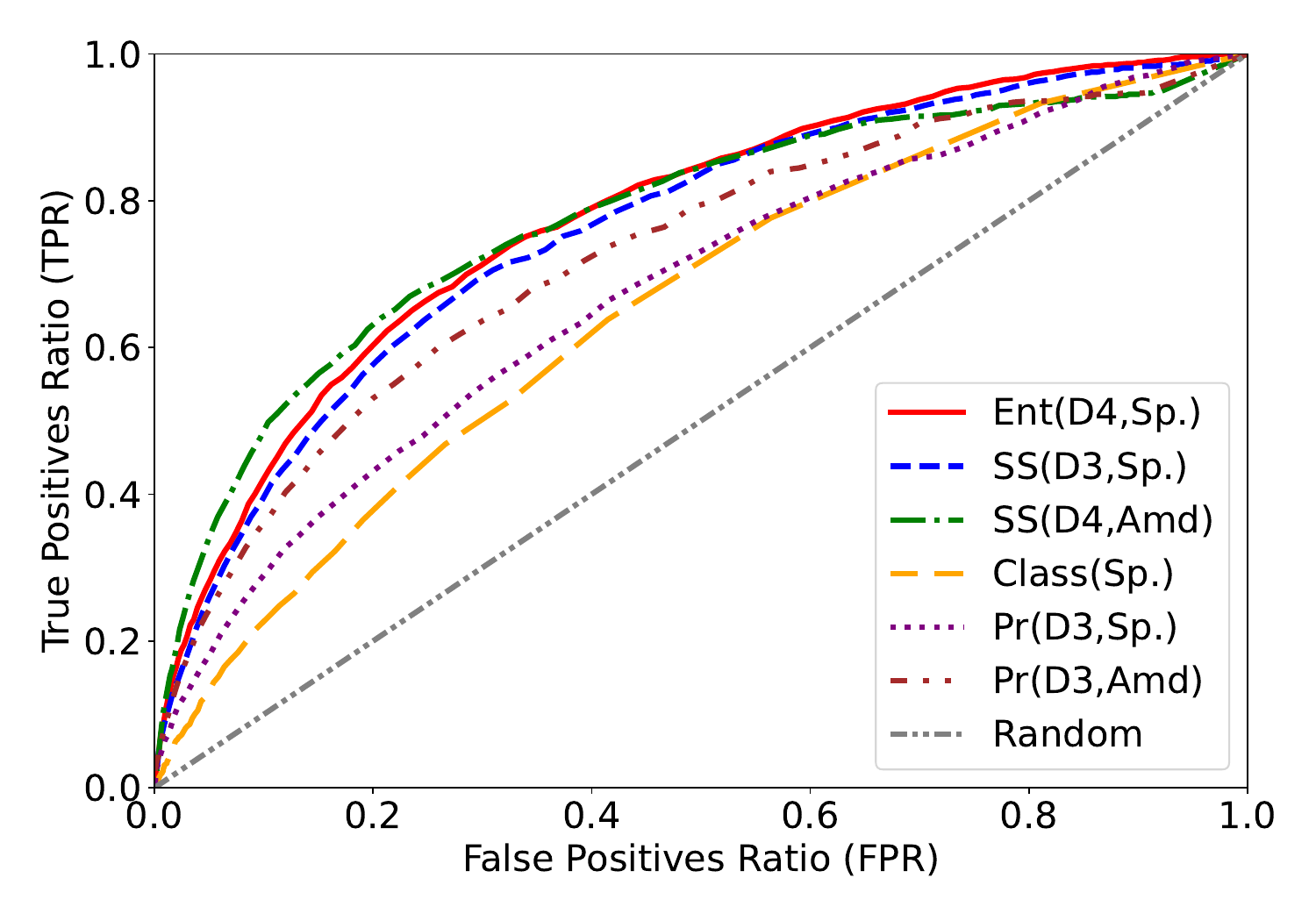} \\
  \includegraphics[width=.9\columnwidth]{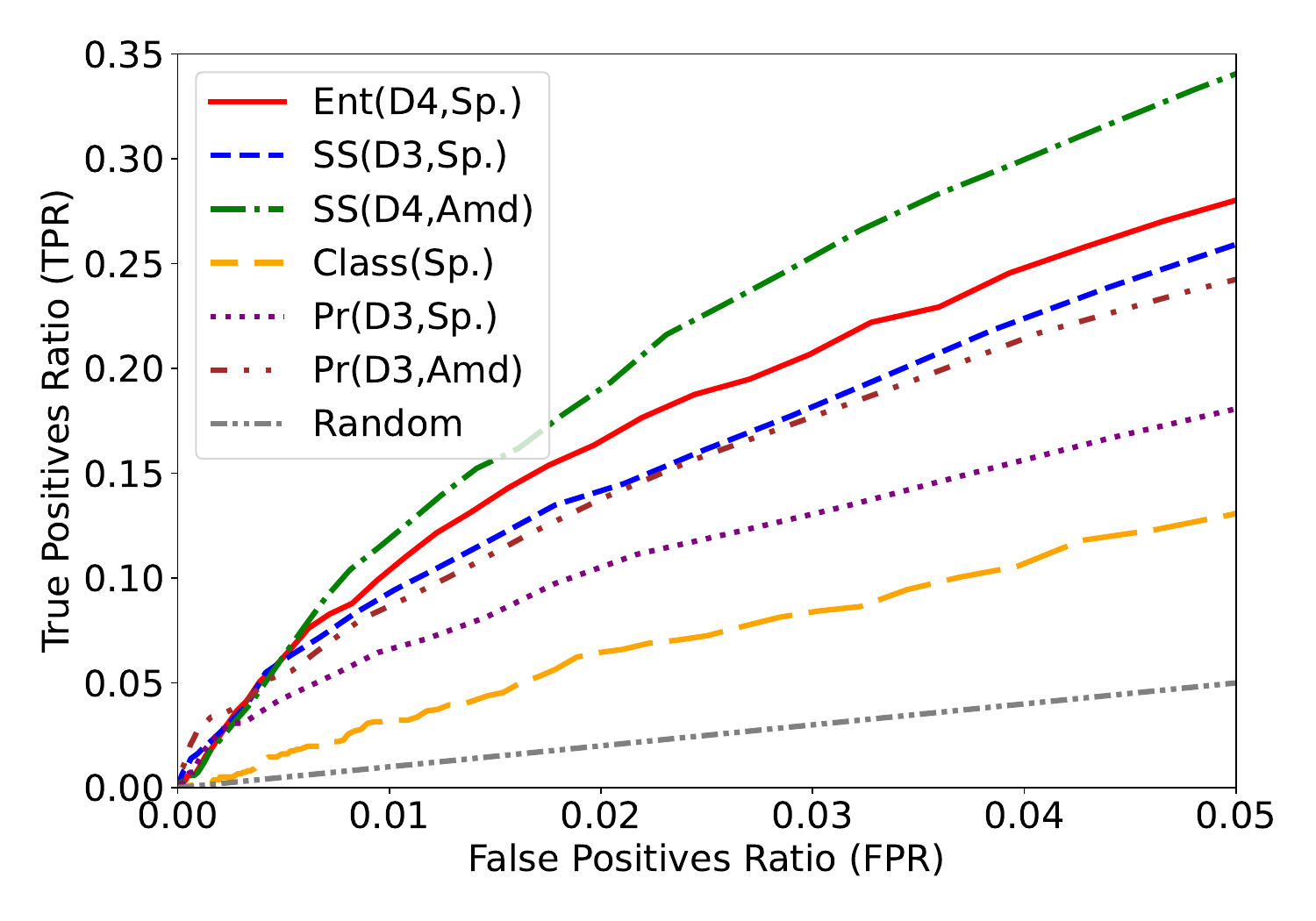}
  \caption{ROC curves for the Meetings dataset - Full (Top) and FPR$\leq$0.05 region (Bottom)}
  \label{fig:meetingsroccurves}
\end{figure}

The scores of all methods are given in Table \ref{tab:scores} and the ROC curves for retweets and meetings are in Figure \ref{fig:twitterroccurves} and Figure \ref{fig:meetingsroccurves} respectively.
We denote in parentheses the documents used for the sets $\mathcal{D}_l$ (D2:all English documents, D3:Position Papers, D4:Summaries) and $\mathcal{S}_m$ (Sp.:Speeches/Speech Summaries, Amd: Amendments).
Methods using Vicuna-generated summaries are indicated by a (V) at the end of the model name; the other methods using summaries use ChatGPT-generated ones.
For a fair comparison between methods, the evaluations include only the set of lobbies that have position papers.

\begin{table}[h]
  \centering
    \begin{tabular}{lcccc}
    \toprule
          \multirow{2}[0]{*}{Method}  & \multicolumn{2}{c}{Retweets} & \multicolumn{2}{c}{Meetings} \\
           & AUC   & pAUC  & \multicolumn{1}{c}{AUC} & \multicolumn{1}{c}{pAUC} \\
    \midrule
    Random & 0.500 & 0.025 & 0.500  & 0.025 \\ 
    Pr(D2,Sp.) & 0.603 & 0.052 & 0.598 & 0.048 \\
    Pr(D3,Sp.) & 0.652 & 0.092 & 0.673 & 0.111 \\
    Pr(D2,Amd) & 0.605 & 0.059 & 0.668 & 0.070 \\
    Pr(D3,Amd) & 0.646 & 0.106 & 0.724 & 0.150 \\
    Nat & 0.530 & 0.076 & 0.551 & 0.107 \\ \midrule  
    Class(Sp.) & 0.687 & 0.079 & 0.652  & 0.070 \\
    SS(D2,Sp.) & \textbf{0.763} & 0.189 & 0.751 & 0.147 \\
    SS(D3,Sp.) & 0.742 & 0.185 & 0.759 & 0.156 \\
    SS(D4,Sp.) (V) & 0.756 & 0.184 & 0.770 & 0.170 \\
    SS(D4,Amd) (V) & 0.701 & 0.153 & 0.766 & 0.198 \\
    SS(D4,Sp.) & 0.759 & 0.196 & \textbf{0.780} & 0.176 \\
    SS(D4,Amd) & 0.704 & 0.169 & 0.773 & \textbf{0.208} \\
    Ent(D4,Sp.) & 0.758 & \textbf{0.198} & 0.774 & 0.175 \\
    \bottomrule
    \end{tabular}%
  \caption{Evaluation results of baselines (top) and our methods (bottom). The pAUC is computed on the region where FPR $\leq 0.05$.}
  \label{tab:scores}%
\end{table}%

We clearly see that the text models that use semantic similarity and entailment outperform all baselines and the text classification model on both datasets.
In fact, the classification model is worse than some baselines. 
We think this could be because it is unable to capture all aspects of a lobby's position in the fixed-length classifier weights, while the similarity-based methods do not have this constraint.

Using only position papers ($D3$) does not seem to have a significant negative effect on performance in the low FPR region compared to using all documents ($D2$).
In fact, for the Prolificacy baselines it results in a significant \textit{increase} in performance.

Summarization seems to help in general for both datasets, especially in the low FPR region.
ChatGPT summaries perform better than Vicuna summaries, with the gap being particularly large for Retweets in the low FPR region.
We, therefore, use ChatGPT summaries for our subsequent entailment method\footnote{The entailment method is computationally expensive to run because of the cross encoder. Therefore we run it only once for the best combination of data (ChatGPT-generated summaries of speeches and lobby documents).}.
It is worth noting however that Vicuna, being free and open-source, is a promising and cheap alternative to ChatGPT.

It is interesting that the model using amendments performs the best for the meetings, whereas it is worse than the model using speeches in the case of retweets.
This could be because amendments are relatively less accessible to the public than speeches, hence this might reflect links that might not be evident in retweets that are very public.
But these links could appear in the meetings data that are relatively less public than retweets.

The entailment method is the best method for the retweet data in the low FPR region and also performs reasonably well in the other cases.
Therefore, we use this method for interpretation.
Although the improvement over semantic similarity in terms of pAUC is small, entailment significantly improves interpretability by reducing false positive matches in the document pairs, as we show in Section \ref{sec:examplematches}.

\section{Interpretation}
\label{sec:lb_interpretation}
We now interpret the links discovered using the entailment method to see if we can find interesting patterns.
To obtain the discovered links, we set the threshold on $A(m,l)$ to 0.7, which gives an FPR of 5\% and TPR of 32.5\% on the Retweets data.
We also manually check a small sample of matched texts and verify that the threshold indeed gives reasonable matches with only a few false positives.

We look at the issues that lobbies are mostly interested in, at their level of focus toward different political groups and ideologies, and at some examples of matched texts that show the method's interpretability.

\subsection{Lobbies and Debates}
\label{sec:lobdebates}
We first look at the issues that lobbies are most interested in by ranking the debates, based on the number of links to lobby clusters and after normalizing by the number of speeches in the debate.
The list of top-five and bottom-five debates is given in Table \ref{tab:topbottomdebates}.

\begin{table}[h]
\centering
\caption{Most and least lobbied debates}
\label{tab:topbottomdebates}
\begin{tabular}{@{}c@{}}
\toprule
\multicolumn{1}{c}{\textbf{Most Lobbied Debates}}       \\ \midrule
European Accessibility Act      \\
Packaging and packaging waste, WEEE \\
Energy efficiency     \\
Plastics in a circular economy  \\
Circular Economy package        \\ \toprule
\multicolumn{1}{c}{\textbf{Least Lobbied Debates}}               \\ \midrule
ESIF: specific measures for Greece \\
Death penalty in Indonesia                              \\
EU-Australia Framework Agreement                        \\
Situation in Iraq                                       \\
Envisaged EU-Mexico PNR agreement                       \\ \bottomrule
\end{tabular}
\end{table}

We see that the most lobbied debates are related to energy efficiency and environmental issues (particularly plastic waste, recycling, and the circular economy), whereas the least lobbied debates are related to international agreements and humanitarian issues.
The apparent lack of lobbying on international issues could be due to the fact that the governments of countries outside the EU are not required to be registered in the EU TR, hence their lobbying activity is not included in our data.

Similarly, we look at the top debates for specific lobby clusters, and these debates make intuitive sense given the area of interest of the lobby.
We give in Table \ref{tab:topdebatesmanufacturing} the top debates for the \textit{Manufacturing} lobby cluster.

\begin{table}[h]
\centering
\caption{Top debates for the \textit{Manufacturing} cluster}
\label{tab:topdebatesmanufacturing}
\begin{tabular}{@{}c@{}}
\toprule
EU-Korea Free Trade Agreement      \\
European Defence Industrial Development \\
Anti-dumping, EU steel industry     \\
Common Commercial Policy \\
Foreign investments in strategic sectors \\ \bottomrule
\end{tabular}
\end{table}

\subsection{Lobbies and Political Groups}
\label{sec:lobpolgroups}
To evaluate the level of focus for a lobby $l$ towards a particular political group $p$, we calculate the \textit{lobby focus score}

\begin{equation}
f(l,p) = \frac{n(l,p)}{m_p}, 
\end{equation}

where $n(l,p)$ is number of discovered links between $l$ and MEPs in $p$, and $m_p$ is the number of MEPs in $p$.
To have comparable scores independent of the size of the lobby, we further normalize them as $\hat{f}(l,p) = \frac{f(l,p)}{\max_{p \in \mathcal{P}} f(l,p)}$ where $\mathcal{P}$ is the set of all 9 political groups.
We analyze at the level of lobby clusters by averaging $\hat{f}(l,p)$ for all the lobbies $l$ in a particular cluster.

\begin{figure}[h]
    \centering
    \includegraphics[trim=0cm 0cm 0cm 0cm, clip, width=0.9\columnwidth]{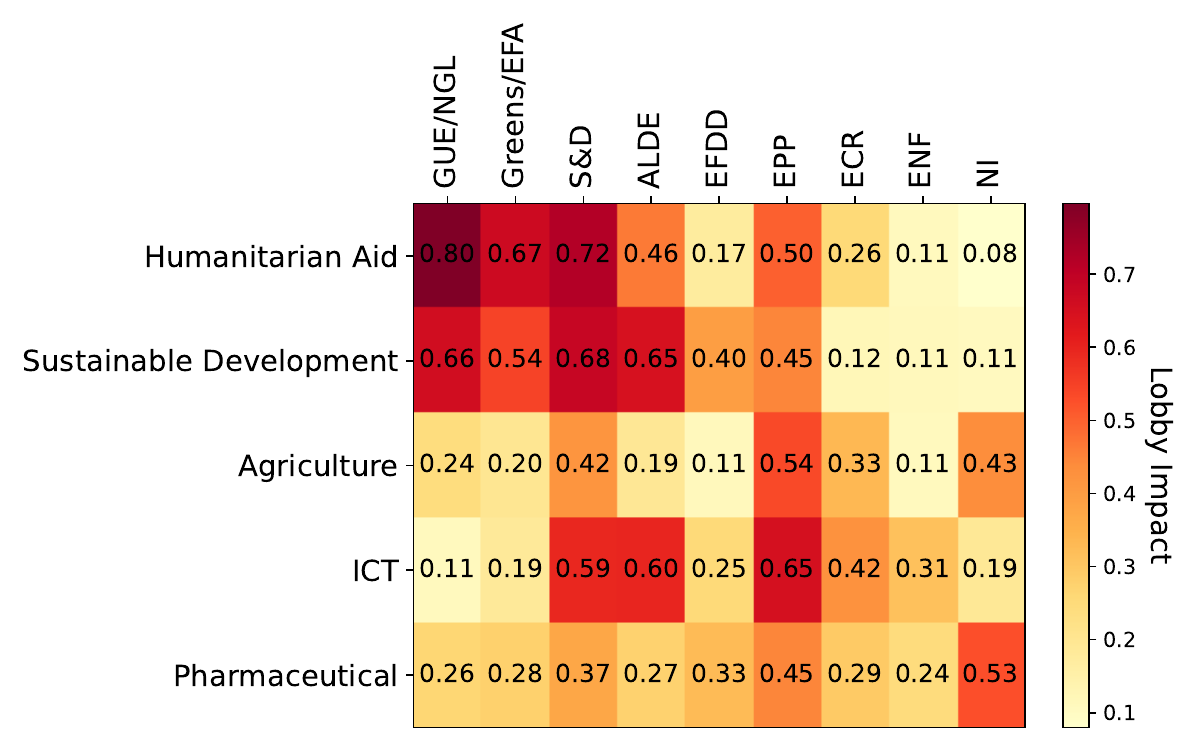}
    \caption{Lobby focus heatmap. Political groups are ordered by ideology from left to right.}
    \label{fig:lobby_heatmap}
\end{figure}

A lobby focus heatmap for selected lobby clusters is given in Figure \ref{fig:lobby_heatmap}.
The political groups are ordered in terms of ideology from left to right.
We see that lobbies associated with social causes and the environment focus on left-leaning groups, whereas agriculture, ICT, and pharmaceutical lobbies focus more on right-leaning groups.

We also show, in Table \ref{tab:topbottomlobbyclusterslrgenscore}, the left-most and right-most lobby clusters, in terms of the weighted average ideology score of the political groups and with the lobby focus score as the weights.
Again, we see that the social and environmental lobbies are aligned to the left, whereas technology, agriculture, and chemical lobbies are aligned to the right.

\begin{table}[h]
\centering
\caption{Top and bottom lobby clusters by ideology score. The numbers in parentheses correspond to the numbers in Figure \ref{fig:lobby_clusterPCA}.}
\label{tab:topbottomlobbyclusterslrgenscore}
\begin{tabular}{c}
\toprule
\textbf{Left-Most Lobby Clusters} \\
\midrule
Social Economic Interests (1)\\
Humanitarian Aid Groups (2)\\
Sustainable Development Groups (3)\\ 
HIV/AIDS advocacy and support (4)\\
Road safety and transportation advocacy (5)\\ \midrule
\textbf{Right-Most Lobby Clusters} \\
\midrule
Technology advocacy groups (6)\\ 
Agricultural interest groups (7)\\ 
Digital and ICT interest groups (8)\\ 
Pharmaceutical and Chemical Advocacy (9)\\ 
Miscellaneous Technology and Education (10)\\
\bottomrule
\end{tabular}
\end{table}

We construct \textit{focus vectors} for the lobbies 

\begin{equation}
    \mathbf{f_l} = \left[ \hat{f}(l,p) ~~\forall p \in \mathcal{P} \right],
\end{equation}

and obtain the focus vectors for lobby clusters by averaging $\mathbf{f_l}$ for the lobbies $l$ in a cluster.
To study how the lobby clusters are arranged in this space, we project them using Principal Component Analysis (PCA).

To interpret each principal component, we compute its Spearman correlation, with the four different ideology scores from the CHES dataset.
Only the first three principal components have statistically significant correlations (p-value below 0.0001).
The results for these are given in Table \ref{tab:pcideo}.

\begin{table}[h]
\centering
\caption{Spearman correlation of principal components with ideology scores. The values in bold have a p-value below 0.0001. The highest absolute values in each row are marked by asterisk(*).}
\label{tab:pcideo}
\begin{tabular}{@{}lllll@{}}
\toprule
              & \textbf{Ideo}  & \textbf{Econ}   & \textbf{Soc}  & \textbf{EU}     \\ \midrule
\textbf{PC 1} & -0.18          & \textbf{-0.41}  & -0.11         & \textbf{-0.47*} \\
\textbf{PC 2} & -0.15          & \textbf{-0.67*} & 0.02          & \textbf{-0.47}  \\
\textbf{PC 3} & \textbf{0.92*} & \textbf{0.51}   & \textbf{0.91} & \textbf{-0.44}   \\ \bottomrule
\end{tabular}
\end{table}

We see that PC 3 and PC 2 have strong correlations with general left-right ideology and the economic aspect of ideology respectively.
PC 3 also has a strong correlation with the social aspect of ideology.

\begin{figure}[h]
    \centering
    \includegraphics[trim=2.5cm 1cm 3cm 2cm, clip, width=0.9\columnwidth]{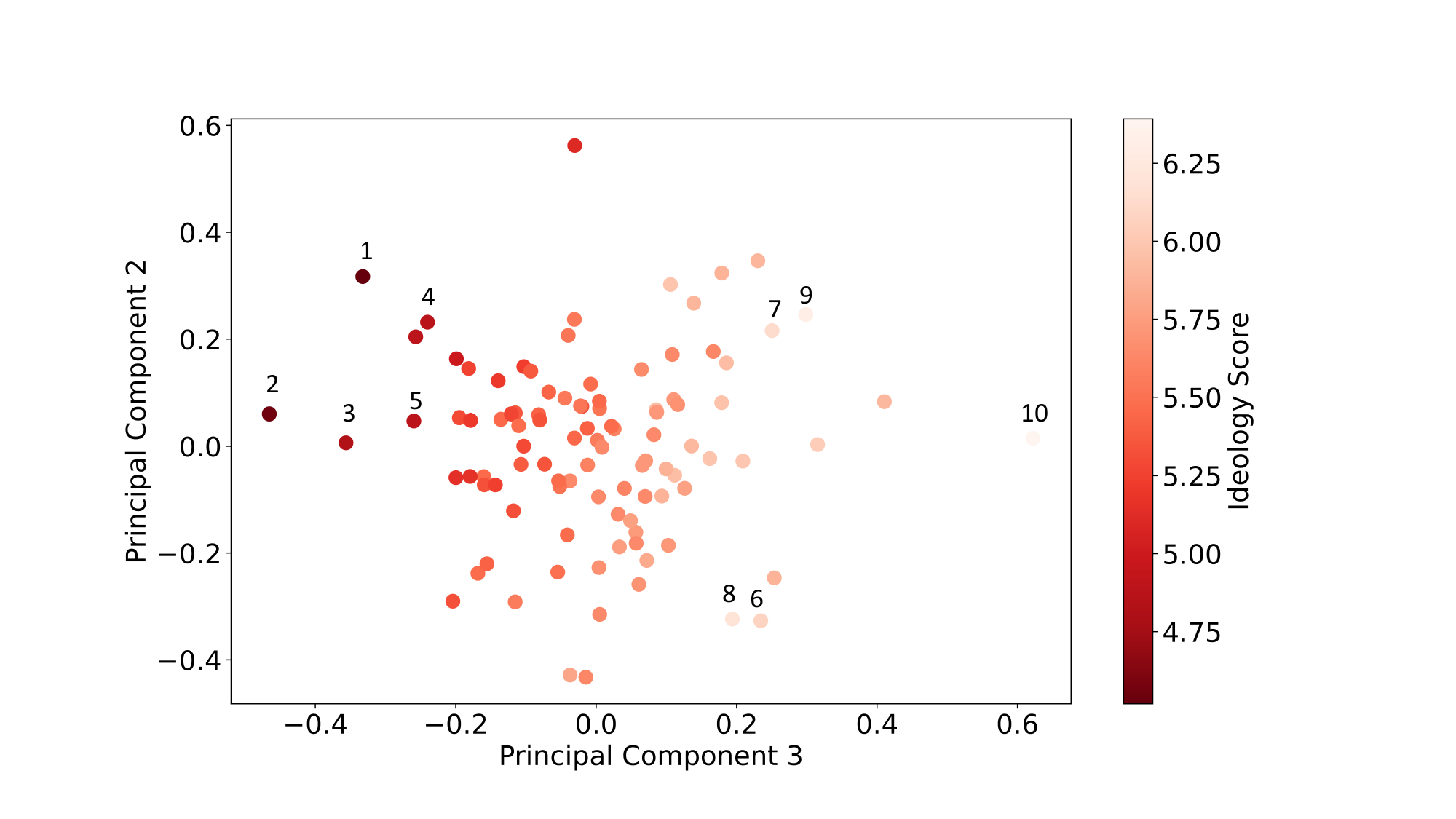}
    \caption{Lobby clusters projected on principal components. The color of the dot corresponds to the general left-right ideology score. The dots annotated with numbers correspond to the clusters in Table \ref{tab:topbottomlobbyclusterslrgenscore}.}
    \label{fig:lobby_clusterPCA}
\end{figure}

To visualize and better understand the lobby clusters, in terms of these ideological dimensions, we project them onto PC 2 and PC 3 and obtain the plot in Figure \ref{fig:lobby_clusterPCA}.

We annotate the dots corresponding to the clusters mentioned in Table \ref{tab:topbottomlobbyclusterslrgenscore}.
In addition to the general left-right placement of these clusters that we already discussed, we also observe their positions with regard to the management of the economy being reflected in the PC 2 coordinates.
In particular, the agriculture lobby (number 7) appears to be in favor of more state control (they are known to be in favor of state subsidies \citep{bednarikova2012agricultural}), whereas the technology lobbies (numbers 6 and 8) appear to advocate for more freedom of the market.

\subsection{Example Matches}
\label{sec:examplematches}
We first look at an example pair of a speech summary and position-paper summary that matched (high semantic similarity and $P(ent) > P(con)$) in Table \ref{tab:examplematch1}.
We see clearly that both documents argue in favor of implementing the Pan-European Pension Product (PEPP) and giving it tax advantages at the national level. 
To demonstrate the advantage of the entailment method, we also show an example pair of a speech summary and position-paper summary that contradict each other (high semantic similarity and $P(con) > P(ent)$) in Table \ref{tab:examplematch2}.
We see that though the speech argues in favor of the EU-US Privacy Shield, the position paper opposes it.
The entailment method is able to avoid such false positives.

\begin{table}[h]
\centering
\caption{Example matching pair of speech summary $s$ and position paper summary $d$. Similar portions of the text are highlighted in bold. $\mathbf{v_s}^T \mathbf{v_d} = 0.916$, $P_{(s,d)}(ent) > P_{(s,d)}(con)$.}
\label{tab:examplematch1}
\begin{tabular}{|p{0.9\columnwidth}|}
\multicolumn{1}{c}{\textbf{Speech Summary}} \\
\midrule
We fully \textbf{support the implementation of the Pan-European Personal Pension Product (PEPP)} \dots We urge Member States to \textbf{grant PEPPs the same tax advantages as similar national products}, \dots \\
\midrule
\multicolumn{1}{c}{\textbf{Position Paper Summary}} \\
\midrule
As an interest group operating in the European Parliament, we believe that the \textbf{Pan-European Personal Pension Product (PEPP) presents an opportunity} \dots making the PEPP simple and transparent, and \textbf{addressing national tax incentives}. Ultimately, making the PEPP a mass-market product remains challenging, and \textbf{tax incentives are crucial to achieve this goal.} \\
\bottomrule
\end{tabular}
\end{table}

\begin{table}[h]
\centering
\caption{Example contradicting pair of speech summary $s$ and position paper summary $d$. Contradicting portions of the text are highlighted in bold. $\mathbf{v_s}^T \mathbf{v_d} = 0.904$, $P_{(s,d)}(con) > P_{(s,d)}(ent)$.}
\label{tab:examplematch2}
\begin{tabular}{|p{0.9\columnwidth}|}
\multicolumn{1}{c}{\textbf{Speech Summary}} \\
\midrule
We strongly support the importance of transatlantic data transmission for our economy, security, and trade. \textbf{The Privacy Shield is a significant step towards achieving much-needed data protection for EU citizens,} \dots to avoid legal uncertainty for our companies and SMEs. \textbf{It is crucial to have an operational Privacy Shield as soon as possible} for the benefit of our companies, the European economy, and the privacy of EU citizens. \\
\midrule
\multicolumn{1}{c}{\textbf{Position Paper Summary}} \\
\midrule
As an interest group operating in the European Parliament, \textbf{we have serious concerns about the proposed EU-U.S. Privacy Shield,} which aims to replace the Safe Harbour framework for commercial data flows between the EU and the U.S. \textbf{We are urging the European Commission not to adopt the Privacy Shield,} as it does not provide adequate protection \dots \\
\bottomrule
\end{tabular}
\end{table}

\section{Summary}
\label{sec:lb_conclusion}
In this paper, we presented an NLP-based approach for discovering interpretable links between MEPs and lobbies, and we collected novel datasets of position papers, speeches, amendments, tweets, and meetings in the process. 
We discovered links that were validated indirectly by using tweets and meetings. 
An aggregate qualitative analysis of discovered links follows expected lines of ideology and the discovered text matches are interpretable. 
We believe our work will help political scientists, journalists, and transparency activists to have a more efficient and larger-scale investigation of the complex links between interest groups and elected representatives. 

\paragraph{Data Limitations}  The Transparency Register is voluntary for several categories of lobby groups, including public authorities of third countries. We could not also include individual companies that are not part of associations, as position papers are difficult to obtain for them.

\paragraph{Methodology Limitations}  We considered only English-language lobby documents. 
There could be some loss of information in the automatic translation of speeches. 
We could summarise only a limited number of lobby documents due to the cost constraints of using ChatGPT (monetary cost) and Vicuna (computational cost). 
With a larger computational budget, larger LLMs such as Vicuna-13B could be tried which could give better performance.

\paragraph{Release of Data}  All data is collected from publically available sources. 
We release data to enable reproducibility while respecting copyright. 
The speeches of the MEPs are made publically available by the EP, and their use and reproduction are authorized. For lobby documents, we do not release copies of the original documents.
We release only the GPT-generated summaries and the URLs of the original documents.
To mitigate link rot, we also release, where possible, links to the archived versions of the documents on the Internet Archive. 
We ensure that the summaries of position papers that we release do not contain any personal data. 
Twitter data is collected through their official API and, following their terms of service, we release only the tweet IDs and not the content or metadata of the tweets.

\paragraph{Possible negative consequences}  
As mentioned in the Introduction, our discovered links only indicate a potential convergence of views between MEPs and lobbies on the issues referenced by the matched texts.
Interpreting them as influence without performing additional investigations could cause harm to MEPs' reputations.
In this paper, we only discussed the results of aggregate analyses to avoid such harm.
Another possible negative outcome of this work is that it could provide hints to some lobby organizations (whose objectives may be against that of the wider society) regarding which political groups they should focus their efforts on.
While this is true, we believe it is important to have transparency regarding this aspect so that the public can be aware of these interactions and be alert to the effects of such influence.

\bibliography{aaai22}

\appendix
\section{Lobby Clusters}

\begin{table}[h]
\centering
\caption{Lobby clusters we refer to in this work and some representative lobbies.}
\label{tab:fulltoplobbyclusters}
\begin{tabular}{@{}ll@{}}
\toprule
\textbf{Lobby Cluster} & \textbf{Example Lobbies}                                                      \\ \midrule
Manufacturing          & \begin{tabular}[c]{@{}l@{}} orgalim.eu \\ glassforeurope.com \end{tabular}       \\ \midrule
Renewable Energy      & \begin{tabular}[c]{@{}l@{}} solarpowereurope.org\\ windeurope.org\end{tabular} \\ \midrule
Business               & \begin{tabular}[c]{@{}l@{}}enterprisealliance.eu\\ smeeurope.eu\end{tabular} \\ \midrule
Social Economic Interests &  \begin{tabular}[c]{@{}l@{}} socialfinance.org.uk \\ nesst.org \end{tabular} \\  \midrule
Humanitarian Aid  & \begin{tabular}[c]{@{}l@{}} ifrc.org \\ voiceeu.org \end{tabular} \\ \midrule
Sustainable Development &  \begin{tabular}[c]{@{}l@{}} milieudefensie.nl \\ zero.ong \end{tabular} \\ \midrule
HIV/AIDS advocacy  &  \begin{tabular}[c]{@{}l@{}} hivjustice.net \\ eatg.org  \end{tabular}  \\ \midrule
Road safety & \begin{tabular}[c]{@{}l@{}} fevr.org \\ eurorap.org  \end{tabular} \\ \midrule
Technology advocacy & \begin{tabular}[c]{@{}l@{}} ecommerce-europe.eu \\ blockchain4europe.eu  \end{tabular} \\ \midrule
Agriculture & \begin{tabular}[c]{@{}l@{}} eurofoiegras.com \\ agricord.org  \end{tabular} \\ \midrule
Digital and ICT & \begin{tabular}[c]{@{}l@{}} all-digital.org \\ digitaleurope.org  \end{tabular} \\ \midrule
Pharmaceutical  & \begin{tabular}[c]{@{}l@{}} medicinesforeurope.com \\ eipg.eu  \end{tabular} \\ \midrule
Miscellaneous Technology & \begin{tabular}[c]{@{}l@{}} claire-ai.org \\ feam.eu  \end{tabular} \\ \bottomrule
\end{tabular}
\end{table}

\end{document}